\documentclass[10pt,letterpaper]{article}

\usepackage{cogsci}
\cogscifinalcopy
\usepackage{pslatex}
\usepackage{apacite}
\usepackage{booktabs} % for professional tables
\usepackage{caption}
\usepackage{subcaption}
\usepackage{graphicx}
\usepackage{amsmath}
\usepackage{amssymb}
\usepackage{mathtools}
\usepackage{amsthm}
\usepackage{float} 
\usepackage{bbm}
\usepackage{bm}
\usepackage{lipsum}

\title{Analyzing the Benefits of Prototypes for Semi-Supervised Category Learning}

 \author{{\large \bf Liyi Zhang (zhang.liyi@princeton.edu)} \\
  Department of Computer Science \\
  Princeton, NJ 08540 USA
\AND {\large \bf Logan Nelson (logann@princeton.edu)} \\
  Department of Psychology \\
  Princeton, NJ 08540 USA
\AND {\large \bf Thomas L. Griffiths (tomg@princeton.edu)} \\
  Departments of Psychology and Computer Science \\
  Princeton, NJ 08540 USA}

\begin{document}

\maketitle

\begin{abstract}

Categories can be represented at different levels of abstraction, from prototypes focused on the most typical members to remembering all observed exemplars of the category. These representations have been explored in the context of supervised learning, where stimuli are presented with known category labels. We examine the benefits of prototype-based representations in a less-studied domain: semi-supervised learning, where agents must form unsupervised representations of stimuli before receiving category labels. We study this problem in a Bayesian unsupervised learning model called a variational auto-encoder, and we draw on recent advances in machine learning to implement a prior that encourages the model to use abstract prototypes to represent data. We apply this approach to image datasets and show that forming prototypes can improve semi-supervised category learning. Additionally, we study the latent embeddings of the models and show that these prototypes allow the models to form clustered representations without supervision, contributing to their success in downstream categorization performance.

% particularly in circumstances favoring prototype models (e.g., well clustered categories, simple boundaries). Our findings show that forming
% prototypes via unsupervised learning can facilitate later category learning.

% These results build on studies of supervised category learning, extending our understanding of the circumstances under which abstract prototype representations are beneficial.
% These results extend the benefits of abstract prototypes beyond the supervised learning literature.

% %hese results build on studies of
% supervised category learning, 

\end{abstract}

\section{Introduction}

%problem
A basic question in cognitive science is when it makes sense to use discrete abstractions to represent a continuous world. Such abstractions have their advantages -- for example, they reduce demands on memory -- but they also potentially discard important information about the original stimulus. One manifestation of the tension between abstraction and faithfulness to the stimulus is in the categorization literature, where prototype models assume that people find abstract representations of categories while exemplar models assume that those categories are represented by remembering labeled examples of the category members themselves. %\blfootnote{Accepted at \textit{Cognitive Science Society} 2024.}

%challenge
The literature on human category learning has extensively explored the circumstances where prototype and exemplar representations make theoretical sense to use and the circumstances under which human learners seem to use them. However, this literature has focused on supervised category learning tasks, where examples are explicitly labeled according to their category membership (although see \citeNP{Fried_Holyoak_1984}). Empirical evaluation of these models has also tended to focus on simplified laboratory stimuli (with some notable exceptions, e.g. \citeNP{Battleday_Peterson_Griffiths_2020, Sanders_Nosofsky_2020}).

%solution
In this paper, we draw on recent advances in machine learning to explore the benefits of prototype representations for semi-supervised category learning with naturalistic images. Our approach is one of rational analysis \cite{Anderson_1990, Anderson_1991}, where we consider the abstract computational problem that human minds have to solve and then evaluate different strategies for solving that problem. We formulate the semi-supervised category learning problem in terms of a learner seeking to find a representation of a set of stimuli that supports subsequent categorization when category labels become available. In this setting, we can define probabilistic  models that make different assumptions about the nature of this representation. By evaluating the consequences of these assumptions for the categorization of naturalistic images, we can assess whether the kind of abstraction assumed in a prototype model helps to support semi-supervised category learning.

%implications
Our results show that merely making the assumption that the world is made up of discrete prototypes increases the quality of representations produced by rational agents, and can improve their performance in later categorization tasks. These results complement previous findings on supervised category learning, helping to create a more complete picture of the settings where abstraction may be beneficial. In particular, our experiments suggest that prototypes improve model categorization by forming clustered representations; the degree to which these representations improve performance depends on the complexity of the dataset's class boundaries.

\section{Models of Categorization}

\subsection{Prototype Theory}

Prototype models are a classic model of categorization, inspired in part by the results of \citeA{Posner1968} and \citeA{Rosch1973-ROSNC}. The prototype of a category is a summary representation with the same structure as a category member, having features that are frequent, average, or otherwise representative \citeA{murphy2003}. New stimuli are then assigned to the category with the closest prototype \cite{ReedPatternRA}.

\subsection{Exemplar Models}

Exemplar models offer an alternative account of how humans categorize. Rather than abstract prototypes, they represent a category simply in terms of the labeled examples of category members \cite{medin1978,nosofsky1986}. Categorization of a new example is then performed by comparing that example with examples retrieved from memory, where the weight assigned to the category label of each exemplar depends on its similarities to the example under consideration. This approach has proven extremely effective in explaining category learning behavior in laboratory experiments.

\subsection{Rational Analysis}

Rational analysis provides a way to understand human behavior by considering the computational problems that human minds solve. In categorization, the problem can be formulated as one of deciding which category $c$ an observation $\bm{x}$ belongs to. This problem can be solved through Bayesian inference, calculating the posterior probability distribution $p(c|\bm{x}) \propto p(\bm{x}|c)p(c)$. The key to this calculation is the probability density $p(\bm{x}|c)$, which indicates how likely it is that an observed example would be generated from category $c$. From this perspective, learning a category becomes a matter of estimating this probability density.

Viewing categorization as probability density estimation gives us a different way to understand prototype and exemplar models \cite{Ashby_Alfonso-Reese_1995}. Prototype models
correspond to {\em parametric density estimation}, with the prototype being the parameter of a distribution of a fixed form (such as the mean of a Gaussian). Exemplar models correspond to {\em nonparametric density estimation}, with close connections to methods such as kernel density estimation. It is also possible to define  models that interpolate between these extremes, such as {\em mixture models} which represent each category as the weighted sum of a set of parametric distributions \cite{Griffiths_et_al_2011}. This perspective also clarifies the tradeoffs implicit in different categorization strategies and provides insight into the settings people apply them to
\cite<e.g.,>{Smith_Minda_1998}. %Recently, these rational accounts of supervised category learning have been complemented by information-theoretic analyses that predict the emergence of different strategies in unsupervised category learning (Martinez, 2023). 

\section{Analyzing Semi-Supervised Category Learning}
%%%
Our goal in this paper is to explore the benefits of abstraction in the context of semi-supervised category learning. In this setting, the learner has an opportunity to observe and model the world without supervision, before being given labeled examples that provide information about category membership. Semi-supervised learning has not been extensively studied in the literature on human category learning (although see \citeNP{Fried_Holyoak_1984}). However, it is a common paradigm in machine learning where unlabeled examples are typically far easier to obtain than labeled examples.

Following the approach taken in the machine learning literature, we will analyze semi-supervised category learning as a representation learning problem. Under this approach, the learner given a stimulus $\bm{x}$ seeks to form a representation $\bm{z}$ that is useful for solving downstream tasks such as categorization. The categorization problem will ultimately be solved by applying a categorization strategy to the representation $\bm{z}$ rather than the raw stimulus $\bm{x}$. The question thus becomes one of how we can find effective representations for a given domain.

As for categorization, the problem of inferring a representation for a stimulus can be formulated as one of Bayesian inference. Given $\bm{x}$ we can apply Bayes' rule to calculate a posterior distribution over representations $\bm{z}$, with $p(\bm{z}|\bm{x}) \propto p(\bm{x}|\bm{z}) p(\bm{z})$. This provides an opportunity to explore the impact of abstraction -- and specifically the use of prototypes -- in the way that we define the prior on representations, $p(\bm{z})$. By taking a mixture model for this prior distribution, we can implicitly assume that the world is made up of discrete clusters that might be characterized by a prototype. Our question of how beneficial prototypes are in semi-supervised learning thus becomes one of assessing the impact of such a prior.

One advantage of expressing this problem in a form that aligns with recent work in machine learning is that we can draw on that work to explore this question with naturalistic stimuli. In particular, this formulation of the problem aligns with the structure of variational auto-encoders, a class of probabilistic models for representation learning that can be applied to images. In the remainder of this section, we describe these models in detail, including how we can explore the use of different prior distributions.

\subsection{Variational Auto-Encoders}

The variational auto-encoder \cite<VAE;>{Kingma2014AutoEncodingVB,Rezende2014StochasticBA} is a class of probabilistic generative models capable of modeling high dimensional data. Given data $\boldsymbol{x}$, it learns a latent space $\bm{z}$ that generates data $\bm{x}$ from a likelihood model $p_{\theta}(\bm{x}|\bm{z})$, also called the \textit{decoder}. The likelihood $p_{\theta}(\bm{x}|\bm{z})$ is parameterized by neural networks whose trainable parameters are denoted $\theta$. While a direct goal is to maximize the marginal likelihood $p(\bm{x})$, it is difficult to do so in complex parameterizations due to the curse of dimensionality. Therefore, VAEs apply variational inference to maximize a lower bound,
\begin{align}
    \log p(\bm{x}) &\geq \mathbb{E}_{q_{\phi}(\bm{z}|\bm{x})}[\log p_{\theta}(\bm{x}|\bm{z}) + \log p(\bm{z}) - \log q_{\phi}(\bm{z}|\bm{x})] %\\
 %   &\coloneqq \mathcal{L}(\phi,\theta), \nonumber
\end{align}
where $q_{\phi}(\bm{z}|\bm{x})$ is the variational posterior of latents $\bm{z}$ given $\bm{x}$ and is also called the \textit{decoder}. It is parameterized by a neural network with trainable parameters $\phi$. $p(\bm{x})$ is the prior distribution of latents $\bm{z}$ and is typically a diagonal Gaussian. The objective can additionally be interpreted as maximizing the power of reconstruction via $\log p_{\theta}(\bm{x}|\bm{z})$, while regularizing the model with $\log p(\bm{z}) - \log q_{\phi}(\bm{z}|\bm{x})$, which encourages $q$ to be close to the prior. %In general, VAEs balance expressive power and computational efficiency, and remain a state-of-art method for representation learning.

\subsection{Putting Prototypes in the Prior}

A number of variants on VAEs have been explored in the machine learning literature, including a variant that makes the assumption we want to test here: that the prior $p(\bm{z})$ encodes an expectation that stimuli can be represented in terms of a set of discrete prototypes. This can be done by defining this prior to be a mixture distribution, with 
\begin{align}
    p_{\lambda}(\bm{z}) = \frac{1}{K}\sum_{k=1}^K q_{\phi}(\bm{z}|\bm{u}_k), \label{eq:vamp}
\end{align}
where $K$ is arbitrarily chosen (but smaller than the number of training datapoints), and $\bm{u}_k$ is a prototype. It has the same shape as a training datapoint, but is randomly initialized and is trainable. We use $p_{\lambda}(\bm{z})$ to indicate that the prior distribution is now parameterized and learned during training.

%The VAE with the `variational mixture of posteriors prior', or VampPrior, utilizes a more expressive prior distribution than the VAE does. It is motivated by the observation that certain regions in the latent space, i.e., areas where the prior $p(\bm{z})$ assigns a high probability, evades being mapped to by the encoder $ q_{\phi}(\bm{z}|\bm{x})$ during training, and is thus assigned a low probability by the encoder. This then results in blurry image generation.

%The VampPrior VAE tackles this problem by using a learnable prior which adapts to the encoder and regularizes the model less strongly. The VampPrior is defined as,

Equation \ref{eq:vamp} was used in a variant of the VAE that uses a ``variational mixture of posteriors prior,'' also known as the VampPrior \cite{pmlr-v84-tomczak18a}, referring to the prototypes as ``pseudo-inputs.''
The  VampPrior VAE objective simply switches the prior distribution in the VAE objective,
\begin{align} \hspace{-2mm}
    \log p(\bm{x}) &\geq \mathbb{E}_{q_{\phi}(\bm{z}|\bm{x})}[\log p_{\theta}(\bm{x}|\bm{z}) + \log p_{\lambda}(\bm{z}) - \log q_{\phi}(\bm{z}|\bm{x})] %\\
%    &\coloneqq \mathcal{L}_{\text{Vamp}}(\phi,\theta, \lambda). \nonumber
\end{align}
Empirically, the VampPrior VAE achieves a higher marginal log-likelihood lower bound than the vanilla VAE \cite{pmlr-v84-tomczak18a}. These models make no use of category labels for training or inference.

\subsection{Summary}

Formalizing the problem of semi-supervised category learning in terms of representation learning provides a natural way to explore the benefits of a prototype-based representation. In this formalization, our question reduces to exploring the impact of using a mixture model to define a prior on the latent representation of a set of stimuli. This question can be answered empirically by comparing the representations that are produced by variational autoencoders that make different assumptions about the prior distribution applied to their latent space. In the remainder of the paper, we empirically evaluate the impact of this assumption for naturalistic images.

\section{Evaluating the Benefits of Prototypes}

We answer two questions: 1) whether a model that forms prototypes categorizes better, and 2) how this categorization ability is related to latent representations and human class labels. 
We expect components of the mixture distribution (Eq.~\ref{eq:vamp}) to regularize the model by clustering data around the prototypes in the latent space, supporting better categorization. We additionally test the hypothesis that prototypes have an increased advantage on datasets with simpler category boundaries, as established for supervised categorization \cite{Ashby_Alfonso-Reese_1995,Martínez_2023}. In addition, we highlight a phenomenon unique to the semi-supervised case where a model learns unsupervised abstractions, but they differ from the human-labeled classes.

\subsection{Datasets}
We trained VAEs making different assumptions about category structure on two image datasets: MNIST \cite{mnist} and CIFAR-10 \cite{cifar10}. Each dataset contains 10 categories. MNIST consists of handwritten digits and provides a setting where categories are relatively well-clustered with simple boundaries. CIFAR-10 consists of naturalistic images that should pose a challenge for unsupervised models learning category boundaries. We then test the models on perturbed versions of both datasets to establish where prototypes are at an advantage.

% \subsection{Qualitative Approach}
% First, we qualitatively compare categorization-awareness in vanilla VAEs and VampPrior VAEs by applying dimensionality reduction and visualization to their learned latent spaces. To do so, for each model we conduct the following steps,
% \begin{enumerate}
%     \item Pass all test datapoints $\bm{x}$ to the decoder: $\bm{z} \sim q_{\lambda}(\bm{z}|\bm{x})$, generating an equivalent number of embeddings $\bm{z}$.
%     \item Use the t-SNE algorithm \cite{JMLR:v9:vandermaaten08a} to reduce the embeddings $\bm{z}$ to two-dimensions each.
%     \item Visualize the dimension-reduced embeddings where colors separate the 10 digit-labels that correspond to these test datapoints.
% \end{enumerate}

\begin{figure*}[h!]
    \centering
    \begin{subfigure}[b]{0.23\textwidth}
        \centering
        \includegraphics[width=\textwidth]{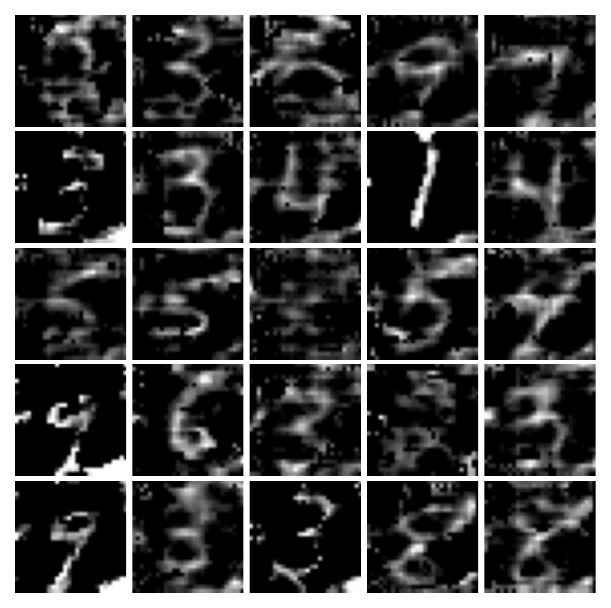}
        \caption{Prototypes (pseudo-inputs).}
        \label{pseudo}
    \end{subfigure}
    \begin{subfigure}[b]{0.23\textwidth}
        \centering
        \includegraphics[width=\textwidth]{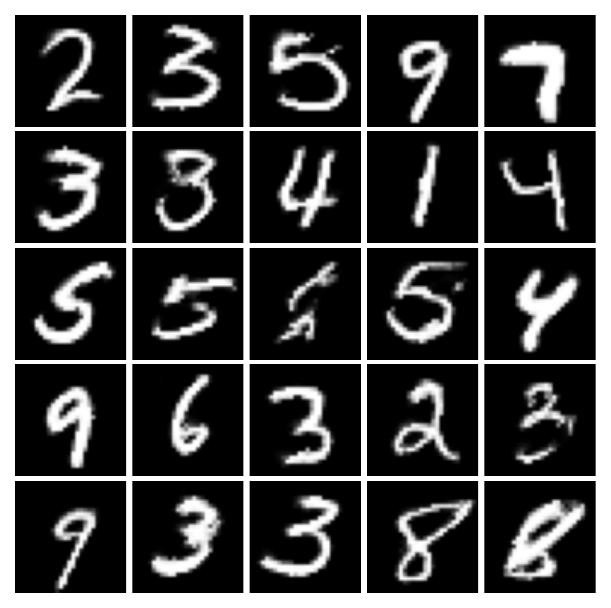}
        \caption{Generations from prior.}
        \label{gen}
    \end{subfigure}
    \begin{subfigure}[b]{0.23\textwidth}
        \centering
        \includegraphics[width=\textwidth]{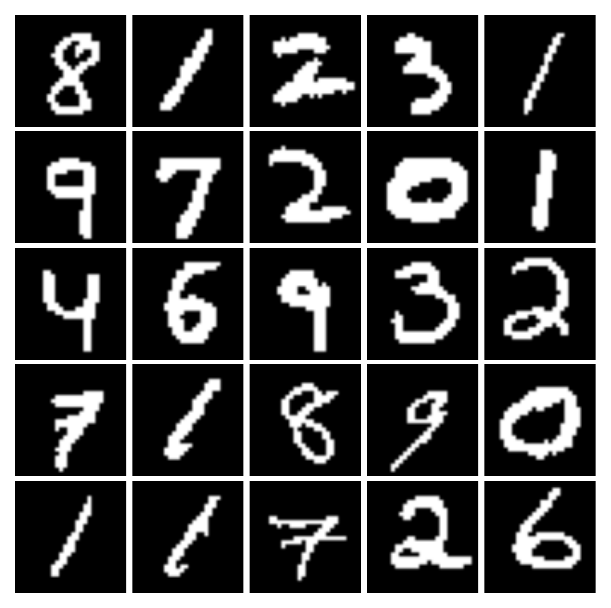}
        \caption{Actual data.}
        \label{dat}
    \end{subfigure}
    \hfill
    \caption{Reproduced results from VampPrior VAE with $K=500$. On the left are the first 25 pseudo-inputs, which are vague versions of real digits. On the middle are generations from these pseudo-inputs (each generated by one pseudo-input). On the right is a random sample of the actual data. The classifier $f$ predicts generated images in the middle as: 2, 3, 5, 9, 7 | 3, 3, 4, 1, 4 | 5, 5, 5, 5, 4 | 9, 6, 3, 2, 3 | 9, 3, 3, 8, 8; the predictions are reasonable even though the generations may be different from the actual data that classifier $f$ was trained on.}
    \label{fig:reproduce}
\end{figure*}

\subsection{Downstream Categorization}

In these methods, we acquire an embedding, or representation, for each datapoint from each model by passing all test datapoints $\bm{x}$ to the model's encoder: $\bm{z} \sim q_{\lambda}(\bm{z}|\bm{x})$, generating an equivalent number of embeddings $\bm{z}$.

\paragraph{K-nearest neighbor (KNN).} First, we classify a datapoint by comparing its embedding to those of its nearest neighbors with known labels. For each VAE, we fit a KNN model on the representations of training datapoints, and make class predictions on the representations of test datapoints.

\paragraph{Classification using prototypes.} We additionally investigate how well the model's learned representations support categorization by using the components in a way as prototypes are typically used to categorize. Specifically, for each datapoint in the embedding space, we find the nearest component based on sum of squares distance, and classify this datapoint based on the class it corresponds to, i.e., for which it is prototypical. In other words, for the MNIST dataset, if pseudo-input $\bm{u}_i$ is a vague representation of digit `6' and is closer to the current datapoint in the latent space than all the other prototypes are, then we classify this datapoint as `6'.

However, these prototypes may not perfectly represent a class. Even if they do, a well-trained VampPrior VAE could use 500 prototypes on a 60000-large MNIST dataset, and it would be laborious to manually assign a label to each one. To bridge this gap, we first train a separate neural network classifier to attain near-perfect accuracy on the dataset. Then, we use the VAE decoder to generate an image from each prototype, and predict the class of this image using the classifier. This class is a surrogate for the class label corresponding to each prototype. In summary, the approach is,
\begin{enumerate}
    \item Train a neural network classifier $f$ on data $\bm{x}$.
    \item For each component $\bm{u}_k$,
    \begin{enumerate}
        \item Generate latent $\bm{z}_k \sim q_{\phi}(\bm{z}|\bm{u}_k)$.
        \item Generate $\bm{x}_k \sim p_{\theta}(\bm{x}|\bm{z}_k))$. 
    \end{enumerate}
    \item Get labels $\bm{y} = f(\bm{X}_K)$, where $\bm{X}_K = \{\bm{x}_k\}_{1\leq k \leq K}$.
    \item For each datapoint $\bm{x}_i$, 
    \begin{enumerate}
        \item Get its embedding with $\bm{z}_i \sim q_{\phi}(\bm{z}|\bm{x}_i)$.
        \item Find the nearest component $\bm{u}_k$, where $k = \text{argmin}_{k \in 1:K} \lVert \bm{z}_k - \bm{z}_i\rVert$.
        \item Classify $\bm{x}_i$ with class $\bm{y}_k$.
    \end{enumerate}
\end{enumerate}

\subsection{Dataset Perturbations}

To evaluate hypotheses that related the prototype advantage with the simplicity of categorization boundaries, we perturb our datasets to alter the complexity of these boundaries. 

% We test whether the implementation of prototypes helps VampPrior VAE create useful representations for categorization later as compared to VAE. We evaluate our models on two datasets, MNIST and CIFAR-10 \cite{mnist, cifar}. 

%We expect the VampPrior VAE, particularly with lower values of $K$, to benefit from datasets in which categories follow a prototypical structure, with simpler category boundaries and less overlap between categories. Thus, we predict more of a benefit for the 

\vspace{1mm}
\noindent\textbf{Adding noise.} For datasets with relatively simple category boundaries (e.g., images of handwritten digits), prototypes are predicted to have an advantage. We then perturb the data in a way that lessens this advantage and simulates more complex boundaries. To do this, we smooth the boundaries between the distributions for each category. This requires we calculate the distribution over binary pixel inputs for each category. Then, each datapoint receives noise from a different, randomly chosen category. To add the noise, the probability of each pixel's value flipping is proportional to how probable that pixel is to be a different value in the other category. Lastly, a parameter $\epsilon$ weights the flip probabilities to control the amount of smoothing between categories.

\vspace{1mm}
\noindent\textbf{Removing high-entropy categories.} For datasets with overlapping, complex category boundaries (e.g., naturalistic images), prototype-like representations should be at less of an advantage. Additionally, any unsupervised model, with or without prototype-learning, can learn category boundaries that are defined differently from the humans who labelled the dataset. For example, as opposed to modeling the nuanced differences between \textit{deer} and \textit{horses}, it may focus instead on separating images with blue vs. green backgrounds. For both these reasons, we examine removing categories with the most overlapping nature prior to training either model.

To identify complex category boundaries, we use human classification judgments (CIFAR-10H; \citeNP{Battleday_Peterson_Griffiths_2020}). People's confusion between categories gives a proxy for the natural complexity and overlap of these categories. The distributions of human responses to each image yields a measure of entropy for each image, and we calculated the categories with the highest entropy. We then train the models after removing these high-entropy categories, allowing us to apply our models to an inherently complex dataset but with simpler categorization boundaries.

\begin{figure*}[h]
    \centering
    \begin{subfigure}{0.33\textwidth}
        \centering
        \includegraphics[width=\textwidth]{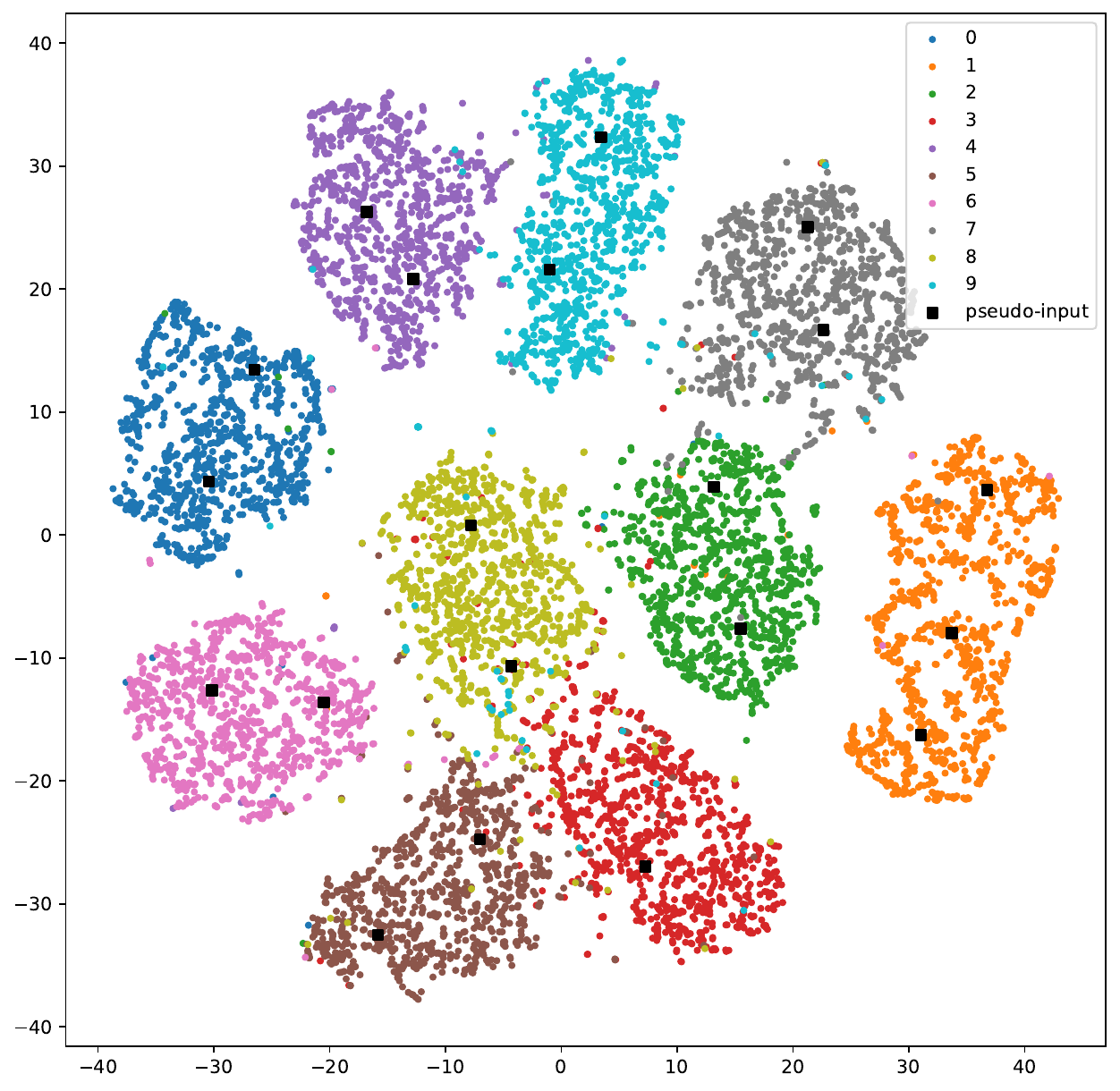}
        \caption{VampPrior VAE with 20 pseudo-inputs.}
        \label{fig:tsne-20}
    \end{subfigure}
    \begin{subfigure}{0.33\textwidth}
        \centering
        \includegraphics[width=\textwidth]{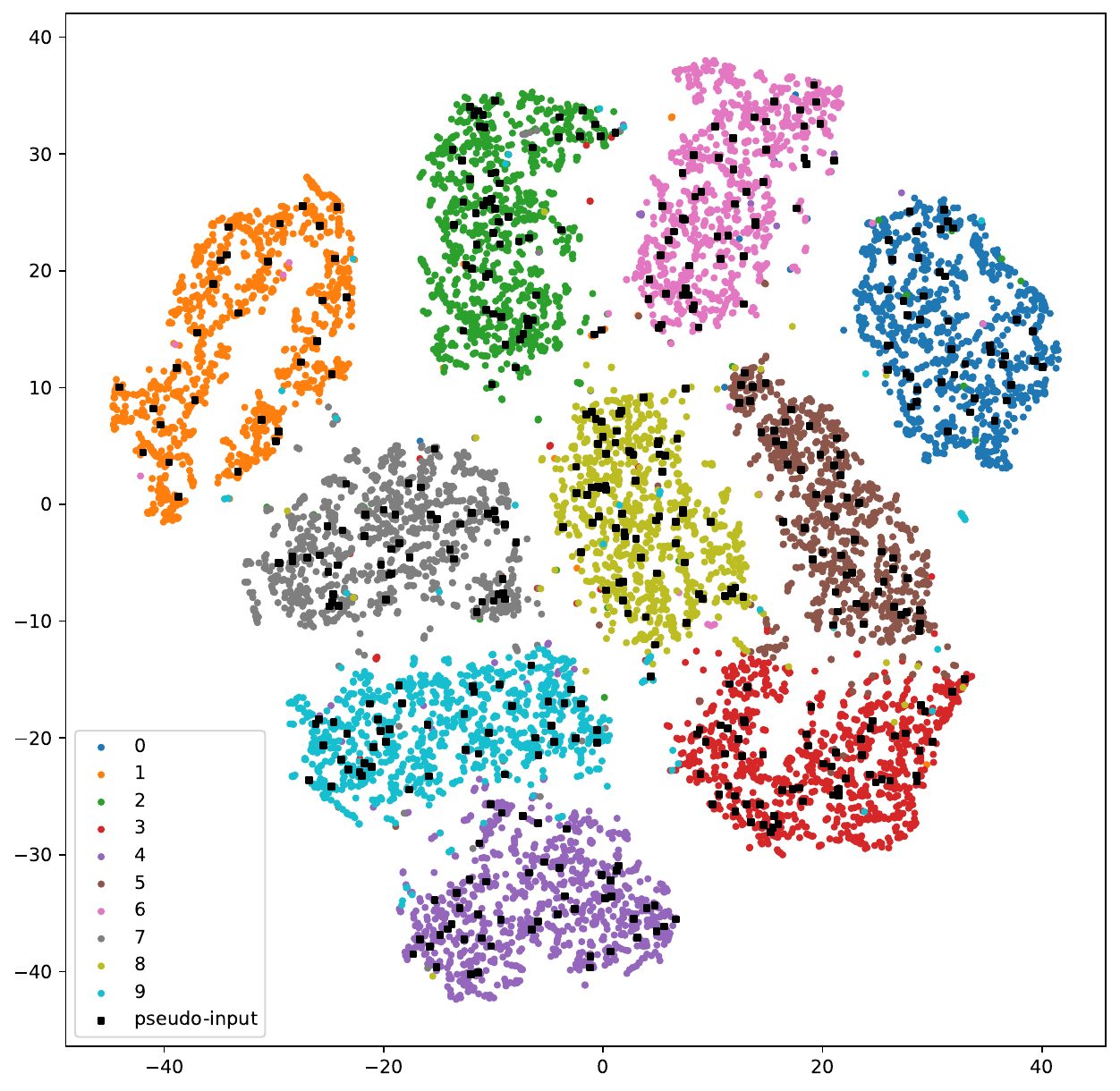}
        \caption{VampPrior VAE with 500 pseudo-inputs.}
        \label{fig:tsne-500}
    \end{subfigure}
    \begin{subfigure}{0.33\textwidth}
        \centering
        \includegraphics[width=\textwidth]{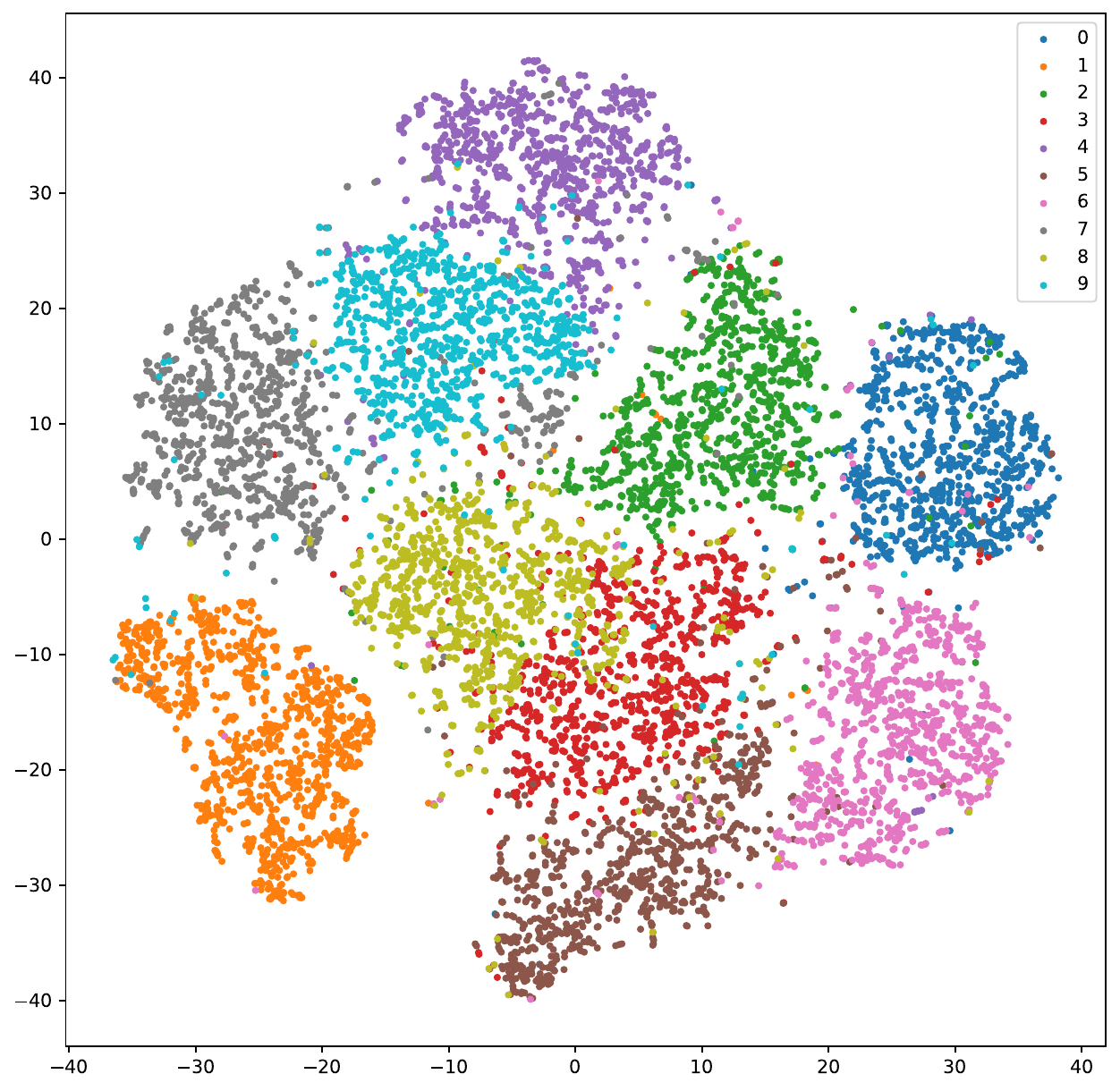}
        \caption{Vanilla VAE.}
        \label{fig:tsne-standard}
    \end{subfigure}
    \caption{t-SNE mapping of test datapoint embeddings for three VAE models. Each colored circle represents a datapoint, and circles are colored by real labels. Black squares are embeddings of pseudo-inputs (prototypes). VampPrior VAE with 20 or 500 pseudo-inputs show stronger demarcation between classes than vanilla VAE, even though no label was present in training.}
    \label{fig:tsne}
\end{figure*}

\subsection{Implementation Details}

To train VAEs on MNIST, we use the original implementation from \citeA{pmlr-v84-tomczak18a} with a two-level VAE using the PixelCNN architecture in encoder and decoder\footnote{Code is available at github.com/zhang-liyi/vampprior-prototype}. Each layer embedding is of size 40. To train them on CIFAR-10, we use a one-level DCGAN architecture in the VAEs \cite{Radford2016UnsupervisedRL}. The layer embedding is of size 200. In all runs, we use learning rate $=0.0005$ and batch size $=100$. The t-SNE algorithm \cite{JMLR:v9:vandermaaten08a} maps embeddings to dimension of size 2 using perplexity $=30$ and number of iterations $=500$. We use a two-layer convolutional neural network with a final linear layer as the separate classifier to assign labels to MNIST prototypes, and it attains over 99\% accuracy. All optimization procedures use the Adam optimizer \cite{Kingma2015AdamAM}.

\section{Results}

\subsection{Reproducing and Understanding VampPrior VAE}

We first illustrate the representations from VampPrior VAE by reproducing a version with $K=500$ on the 60000-large MNIST dataset (Figure \ref{fig:reproduce}). Prototypes from Figure \ref{pseudo} are themselves blurrier versions of the original data. Generations from these prototypes by applying $\bm{z} \sim q_{\phi}(\bm{z}|\bm{u}_k)$ have recognizable relations with these prototypes (Figure \ref{gen}). 

\subsection{Clustered Representations in VAE Embeddings}

We illustrate the latent embeddings that are learned by different VAEs, and show that VAEs that learn prototypes form clustered embeddings.

\paragraph{MNIST.} We visualize the test datapoint embeddings learned by VampPrior VAEs with $K=20$ and $500$, as well as those from the vanilla VAE (Figure \ref{fig:tsne}), where embeddings are dimension-reduced by t-SNE and colored by true digit class. Prototypes are additionally plotted on top of datapoint embeddings. We find that prototypes occupy centric locations among clusters of datapoints. When $K=20$ (Figure \ref{fig:tsne-20}), each of the 10 classes gets at least one prototype, and in each class, prototypes are evenly spaced. Even with $K=500$ (Figure \ref{fig:tsne-500}), prototypes fall within datapoint clusters. For example, digit 1 (orange) is embedded in two close areas, and prototypes evenly fill out the areas, with no prototype falling in between these two areas. Since prototypes are summary representations of categories, these results support the interpretation of VampPrior pseudo-inputs as prototype-based representations.

VampPrior VAE shows clear demarcation between classes, even though the objective function does not lead it to learn classes or clusters. This phenomenon is similar between $K=20$ and $K=500$, but is much weaker for the vanilla VAE (Figure \ref{fig:tsne-standard}). In other words, having the prior $p(\bm{z})$ assume a discrete, prototypical structure of the data allows the model to form understanding of categories in an unsupervised manner. 

\begin{figure}
    \centering
    \includegraphics[scale=0.38]{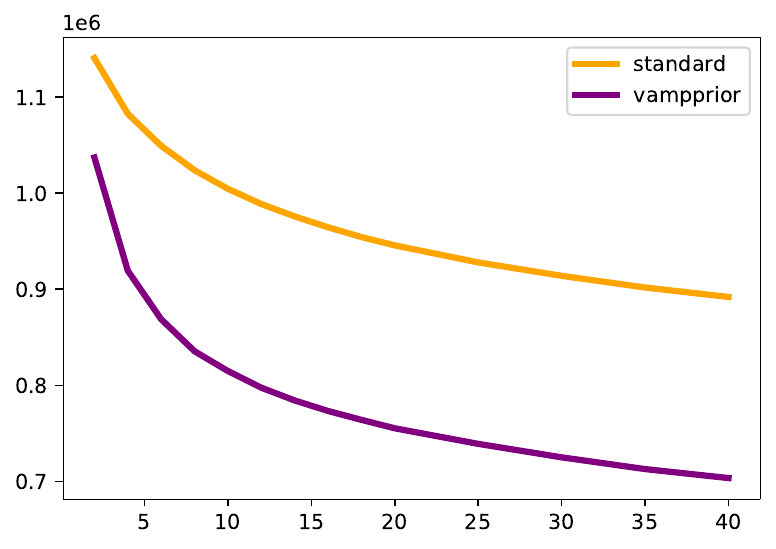}
    \caption{Squared loss from k-means clustering on two VAEs' embeddings, versus the number of clusters used by k-means. \textit{Lower value means more clustered embeddings.}}
    \label{fig:kmeans-cifar10}
\end{figure}

\paragraph{CIFAR10.} CIFAR10 classes are easily confused for an unsupervised model, so we provide a quantitative measure of the degree of clustering in the embeddings learned by VampPrior and standard VAE. We use the k-means clustering algorithm on the test datapoint embeddings from the two VAEs,  scaled by their standard deviations. Then, we use the clustering loss as a measure of how clustered these embeddings are. Figure \ref{fig:kmeans-cifar10} shows that VampPrior VAE embeddings are consistently more clustered than the standard VAE embeddings. 

These studies on VAE embeddings suggest that learning prototypes encourages representations to form clusters in the vector space without supervision. 

\subsection{Classification Results}

\paragraph{Classification using prototypes as reference-points.} 
So far we have studied the patterns that appear in VAE embeddings without making use of real class labels (aside from the colors in the MNIST visualization). Now we examine how the cluster pattern in the embeddings relates to downstream classification performance. We abide by prototype theory: classifying a new datapoint by finding its closest pseudo-input (prototype), and using the prototype's class to predict the class of this datapoint.

Results are shown in Table \ref{tab:acc} with VampPrior VAEs with $K=500,100,20,10$. Accuracy remains high for $K \geq 20$. The lower bound objective decreases as $K$ decreases down from 500, whereas classification accuracy drops from $K=500$ but remains similar for $K=100$ and $20$. This result suggests that an unsupervised learning model implicitly encodes classification capability, and this capability can be recovered from its prototypes. In the extreme case of $K=10$, the 10 prototypes represent only 8 out of the 10 digits, with 2 digits each getting 2 prototypes. Therefore, performance is upper-bounded by $80\%$ accuracy. The generally high accuracy suggests that prototypes help the model differentiate the categories.

\begin{table}[!ht]
\begin{center} 
\caption{Classification accuracy on MNIST by finding the nearest prototype for each datapoint. Closeness is measured by sum of squared distance on the VAE embedding space.} 
\resizebox{0.48\columnwidth}{!}
{
\begin{tabular}{lll} 
\hline
$K$    &  Training Acc & Test Acc \\
\hline
500        &   93.2\%  & 92.5\% \\
100   &   85.1\% & 85.7\% \\
20           &   87\% & 86.6\% \\
10          &   67.4\% & 68\% \\
\hline
\end{tabular}} 
\label{tab:acc}
\end{center} 
\end{table}

\noindent\textbf{Nearest-neighbor classification performance.} Here we quantitatively measure how the different VAEs' representations support categorization by using KNN on these representations (Table \ref{tab:acc-knn}). VampPrior VAE consistently outperforms the vanilla VAE on both MNIST and CIFAR-10. Both VampPrior and vanilla VAEs provide accurate representations for MNIST, but the demarcation in VampPrior's higher performance ($K = 20$ vs.~$10$) corresponds to the place where every digit at least gets one prototype, suggesting prototype's role in boosting the separation of category boundaries. CIFAR-10 results show that higher numbers of prototypes ($K\geq 100$) are able to gain an edge in performance vs.~non-prototype models on this dataset. Given the small edge in performance and the inherent challenge in learning categories for naturalistic images, we study whether the performance gap increases if the most overlapping classes are removed (Table \ref{tab:entropy}).

\begin{table}[!ht]
    \centering
    \caption{Classification accuracy based on the K-nearest-neighbor algorithm (KNN). $K$ refers to the number of prototypes, and $K=0$ refers to the case of the original VAE. }
    \resizebox{0.59\columnwidth}{!}{
\begin{subtable}{0.15\textwidth}
\centering
\begin{tabular}{lll} 
\hline
$K$    &  Test Acc \\
\hline
15000 & 97.8\% \\
10000 & 98\% \\
500        & 98.1\% \\
100   & 97.3\% \\
20            & 97.8\% \\
10          & 94.8\% \\
0 & 94.2\% \\
\hline
\end{tabular}
\caption{MNIST.}
\end{subtable}
\hspace{2em}
\begin{subtable}{0.15\textwidth}
\centering
\begin{tabular}{lll} 
\hline
$K$    &  Test Acc \\
\hline
2000        & 42.2\% \\
1000   & 41.7\% \\
500 & 41.6\%\\
100 & 42\%\\
20  & 41.1\%\\
10  &  40.6\%\\
0 & 41.3\% \\
\hline
\end{tabular}
\caption{CIFAR-10.}
\end{subtable}}
\label{tab:acc-knn}
\end{table}

%\paragraph{Classification accuracy based on the K-nearest-neighbor algorithm (KNN).} Here we wish to study whether the optimal number of prototypes corresponds to the number of datapoints. Tentatively put here (Table \ref{tab:small-data}). It seems that results here (10,000 MNIST) are similar to the 50,000 MNIST situation. 2000 MNIST results in VampPrior VAE (tried K=50 and 500) not learning any meaningful prototypes.

\noindent \textbf{Classification performance and noise levels.} 
The MNIST dataset has relatively well-clustered and simple category boundaries, so we consider a case where prototypes are expected to be at less of an advantage. To do this, we apply our method of smoothing the categories together by adding noise. These results are shown in Table \ref{tab:smoothing}. These results show that the prototype models' performances decrease and converge towards vanilla VAE as smoothing is increased. This is particularly emphasized for lower values of $K$, which have less flexibility due to their fewer prototypes. 

% \begin{table}[!ht]
%     \centering
%     \caption{Classification accuracy on MNIST for different values of the category smoothing parameter $\epsilon$. The case of $\epsilon=0$ refers to no smoothing between categories (hence, these are the same percentages from Table \ref{tab:acc-knn}). In the table, $K$ refers to the number of prototypes as in previous notations, and $K=0$ refers to the case of the original VAE.}
% \begin{tabular}{llllll} 
% \hline
% $K$  &  $\epsilon$=0 &  $\epsilon$=0.3   &  $\epsilon$=0.35  &  $\epsilon$=0.4   & $\epsilon$=0.5\\
% \hline
% 500   & 98.1\% & 98.3\% & 98.0\% & 97.9\% & 97.7\%\\
% 100   & 97.3\% & 97.6\% & 97.4\% & 97.6\% & 96.8\%\\
% 20    & 97.8\% & 96.2\% & 96.5\% & 96.1\% & 94.9\%\\
% 0     & 94.2\% & 95.2\% & 97.1\% & 96.1\% & 97.1\%\\
% \hline
% \end{tabular}
% \label{tab:smoothing}
% \end{table}

\begin{table}[!ht]
    \centering
    \caption{Classification accuracy on MNIST smoothing parameter $\epsilon$. $\epsilon=0$ refers to no smoothing between categories (so these are the same values from Table \ref{tab:acc-knn}). $K$ refers to the number of prototypes, and $K=0$ refers to the original VAE.}
\resizebox{0.9\columnwidth}{!}
{
\begin{tabular}{lllllll} 
\hline
$K$  &  $\epsilon=0$ &  $\epsilon=0.2$   &  $\epsilon=0.3$  &  $\epsilon=0.4$   & $\epsilon=0.5$ & $\epsilon=0.6$\\
\hline
500   & 98.1\% & 97.6\% & 96.3\% & 93.1\% & 82\% & 67.9\% \\
100   & 97.3\% & 97\% & 95.8\% & 92.6\% & 80\% & 65.4\% \\
20    & 97.8\% & 96.2\% & 93.5\% & 89.1\% & 68.1\% & 59.8\% \\
10    & 94.8\% & 94.5\% & 90\% & 87\% & 65.4\% & 62.3\% \\
0     & 94.2\% & 92.4\% & 91.8\% & 86.8\% & 80.3\% & 65.4\% \\
\hline
\end{tabular}}
\label{tab:smoothing}
\end{table}

The CIFAR-10 dataset has more complex boundaries for some of its naturalistic images; humans and an unsupervised learning model may not agree on the way to label this dataset. We removed high-entropy categories based on the mean entropy of responses in human classification performance for each category. Specifically, we conducted two sets of experiments where the top-2 and top-4 most challenging categories are removed. The models are then trained and evaluated on the reduced datasets. The top-4 challenging categories are: deer, cat, bird, and airplane. KNN-based results are shown in Table \ref{tab:entropy}. Random-guess performance would increase from the original 10\% to 12.5\% (top-2 removed) and 16.7\% (top-4 removed). Here, both VAEs demonstrate performance increases, but the outperformance from VampPrior becomes evident as more categories are removed.

\begin{table}[!ht]
    \centering
    \caption{Classification accuracy on CIFAR-10 with removal of high entropy categories. Classification is based on the K-nearest-neighbor algorithm (KNN). $K$ refers to the number of prototypes, and $K=0$ refers to the case of the original VAE.}
\centering
\resizebox{.67\columnwidth}{!}{
\begin{tabular}{llll} 
\hline
$K$  &  Original & 2 Removed & 4 Removed \\
\hline

2000 & 42.2\% & 49.2\% & 55.8\%\\
1000 & 41.7\% & 49.9\% & 56.3\%\\
500  & 41.6\% & 49.4\% & 55.5\%\\
100  & 42\% & 48.9\% & 56.7\%\\
20   & 41.1\% & 48.9\% & 55.8\%\\
10   & 40.6\% & 49.1\% & 55.3\% \\
0    & 41.3\% & 45.9\% & 51.7\%\\
\hline
\end{tabular}}
\label{tab:entropy}
\end{table}

%\paragraph{Nearest-neighbor classification on low-data regime.} Here we wish to study whether the optimal number of prototypes corresponds to the number of datapoints. Tentatively put here (Table \ref{tab:small-data}). It seems that results here (10,000 MNIST) are similar to the 50,000 MNIST situation. 2000 MNIST results in VampPrior VAE (tried K=50 and 500) not learning any meaningful prototypes.

%\begin{table}[!ht]
%    \centering
%    \caption{Classification accuracy based on the K-nearest-neighbor algorithm (KNN) on MNIST with 10,000 training datapoint. In the table, $K$ refers to the number of prototypes as in previous notations, and $K=0$ refers to the case of the original VAE.}
%\begin{tabular}{lll} 
%\hline
%$K$    &  Test Acc \\
%\hline
%500        & 98.3\% \\
%50   & 96.6\% \\
%0 & 93.9\% \\
%\hline
%\end{tabular}
%\label{tab:small-data}
%\end{table}

\section{Discussion} 

How people leverage different kinds of representations, with different levels of abstraction, has been extensively explored in the context of supervised categorization. We show the benefits of prototypes in a semi-supervised setting, where a bias for abstraction in an unsupervised model can enhance its later ability to support categorization.

In line with the supervised categorization literature, our results show that prototypes are beneficial when the category boundaries are simple. This kind of finding is traditionally attributed to prototypes lacking flexibility, but our analyses of the models' latent space suggest a novel factor for semi-supervised learners: an unsupervised learner may form abstractions that differ from those corresponding to human labels. The VampPrior model consistently formed more clustered representations, even on the complex boundaries of the CIFAR-10 dataset. Notably, this clustering behavior is also what drives later improvements in classification.

The results we have presented extend our understanding of the circumstances under which abstract prototype representations are beneficial. Our approach also illustrates how cognitive models of categorization can be implemented in unsupervised and semi-supervised contexts, extending previous work using deep learning to model categorization in naturalistic settings. 

\vspace{2mm}

\noindent {\bf Acknowledgments.} This work was supported by grant number
N00014-23-1-2510 from the Office of Naval Research.

%Although models with different numbers of prototypes had similar classification behavior in our experiments, an open direction is to investigate the properties of having more or less prototypes, with extremely high values potentially behaving more like exemplar models. Another future direction is to further explore the nature of the prototypes learned, and

%what kind of abstractions does it pull out? when you don't have labels, might need more pseudoinputs than classes
%exploring large values of K to perhaps mimic very flexibile exemplar models

\bibliographystyle{apacite}

\setlength{\bibleftmargin}{.125in}
\setlength{\bibindent}{-\bibleftmargin}

\bibliography{main}

% \newpage
% \appendix
% \section{Appendix}

\end{document}